\documentclass[conference]{IEEEtran}
\IEEEoverridecommandlockouts
\usepackage{cite}
\usepackage{amsmath,amssymb,amsfonts}
\usepackage{graphicx}
\usepackage{textcomp}
\usepackage{xcolor}
\usepackage{multirow,algorithm}
\usepackage[algo2e]{algorithm2e}
\usepackage{algorithm,algpseudocode}
\usepackage{subcaption}
\def\BibTeX{{\rm B\kern-.05em{\sc i\kern-.025em b}\kern-.08em
    T\kern-.1667em\lower.7ex\hbox{E}\kern-.125emX}}
\begin{document}

\title{Towards Solving Large-scale Expensive Optimization Problems Efficiently Using Coordinate Descent Algorithm\\
}

\author{Shahryar Rahnamayan, SMIEEE$^1$, Seyed Jalaleddin Mousavirad$^2$  \\
	\ \\
	
	$^1$Nature-Inspired Computational Intelligence (NICI) Lab,\\ Department of Electrical, Computer, and Software Engineering,\\ Ontario Tech University, Oshawa, Canada\\	
	$^2$Faculty of Engineering, Sabzevar University of New Technology, Sabzevar, Iran \\
	 Shahryar.rahnamayan@uoit.ca, Jalalmoosavirad@gmail.com
}

\maketitle

\begin{abstract}
Many real-world problems are categorized as large-scale problems, and metaheuristic algorithms as an alternative method to solve large-scale problem; they need the evaluation of many candidate solutions to tackle them prior to their convergence, which is not affordable for practical applications since the most of them are computationally expensive. In other words, these problems are not only large-scale but also computationally expensive, that makes them very difficult to solve. There is no efficient surrogate model to support large-scale expensive global optimization (LSEGO) problems. As a result, the algorithms should address LSEGO problems using a limited computational budget to be applicable in real-world applications. Coordinate Descent (CD) algorithm is an optimization strategy based on the decomposition of a $n$-dimensional problem into $n$ one-dimensional problem. To the best our knowledge, there is no significant study to assess benchmark functions with various dimensions and landscape properties to investigate CD algorithm and compare with other metaheuristic algorithms. In this paper, we propose a  modified Coordinate Descent algorithm (MCD) to tackle LSEGO problems with a limited computational budget. Our proposed algorithm benefits from two leading steps, namely, finding the region of interest and then shrinkage of the search space by folding it into the half with exponential speed. One of the main advantages of the proposed algorithm is being free of any control parameters, which makes it far from the intricacies of the tuning process. The proposed algorithm is compared with cooperative co-evolution with delta grouping on 20 benchmark functions with dimension 1000. Also, we conducted some experiments on CEC-2017, $D=10, 30, 50$, and $100$, to investigate the behavior of MCD algorithm in lower dimensions. The results show that MCD is beneficial not only in large-scale problems, but also in low-scale optimization problems.   
\end{abstract}

\begin{IEEEkeywords}
Large-scale expensive optimization, Folding search space, Metaheuristics, Coordinate descent
\end{IEEEkeywords}

\section{Introduction}
\label{sec:Intro}
Finding the global optimum of a complicated problem has been one of the long-term goals in the field of computer science and applied mathematics. The problems in real-world applications are usually encountered with some characteristics such as non-linearity, non-convexity, multi-modality, and non-differentiability~\cite{LSGO_benchmark}. 

Numerous problems in real-world applications deal with a large number of decision variables, known as Large-scale Global Optimization (LSGO) problems. Some examples of LSGO problems include large-scale scheduling problems~\cite{LSGO_scheduling_01,LSGO_scheduling_02}, finding weights in a deep belief neural networks~\cite{Evolutionary_Deep}, and large-scale vehicle routing~\cite{LSGO_vehicle_01}. A popular learning model in pattern recognition is multi-layer neural networks (MLNN); the estimation of connection weights in this learning model is generally a challenging task, especially for problems with a high number of input features. For example, for a dataset with 34 features, an MLNN with one hidden layer and 69 neurons in the hidden layer, there are 2,416 connection weights. It arises in deep belief neural networks with more challenges; for instance, an LSTM network with an input size of 4,096 and output size of 256 has 4,457,472 parameters.

In recent years, LSGO is regarded as a well-recognized field of research. To this end, metaheuristic algorithms such as simulated annealing (SA)~\cite{SA_Main_Paper}, particle swarm optimization (PSO)~\cite{PSO_Main_Paper,PSO_Main_Paper02}, differential evolution (DE)~\cite{DE_Original}, and human mental search (HMS)~\cite{HMS_Main_Paper} are extensively employed due to their global search capability and robustness~\cite{GHMS-RCS}. The complexity of LSGO problems and the power of metaheuristic algorithms have been the motivation for many researchers to utilize metaheuristic algorithms for solving LSGO problems. Organizing special sessions in outstanding conferences such as GECCO and CEC, the great number of papers published in decent journals such as IEEE Transactions on Evolutionary Computation, developing new large-scale benchmark functions for competitions, and LSGO-related websites can verify the importance of LSGO research field.

Generally speaking, metaheuristics in the basic form suffer from some shortcomings for solving LSGO problems~\cite{large_scale_survey}, and the performance of them dramatically decreases when dealing with high-dimensional problems~\cite{large_scale_survey,LSGO_ICC,LSGO_CCS}. There are two main reasons for the performance suffering of metaheuristic algorithms in large scale problems: 1) with increasing the number of dimensions, the complexity of the landscape is crucially increased, and 2) the search space grows exponentially. As a result, a metaheuristic algorithm should be able to explore the whole search space effectively, which is not a trivial task especially when the fitness evaluation budget is a small amount~\cite{LSGO_CC01,large_scale_survey}.

From the literature, two main categories of approaches can be observed to tackle LSGO problems, namely problem decomposition strategy, which is based on cooperative co-evolution (CC) algorithm~\cite{CC_survey,CC_main_paper01,CC_main_paper02} and non-decomposition approaches. Non-decomposition approaches solve LSGO problems as a whole which they are designed with some specific operators to explore the whole search space effectively ~\cite{center_sampling_02,center_sampling_05,LSGO_PSO01}. The decomposition methods benefit from a divide-and-conquer approach to decompose an LSGO problem into low dimensional subcomponents~\cite{CC_survey}.

One of the main problems to tackle LSGO problems is that they require evaluation of numerous candidate solutions before convergence that may not be affordable for practical applications that include expensive computational evaluations, such as aerodynamic design optimization~\cite{LSGO_Expensive_01}, drug design~\cite{LSGO_Expensive_02}, and flow-shop scheduling problems~\cite{LSGO_Expensive_03}. As mentioned in~\cite{LSGO_Expensive_03}, it takes 200 days to solve a problem with ten jobs and five machines, which is impractical. As a result, the algorithms need to move towards solving LSGO problems with a limited computational budget so that they can be used for real-world applications. In other words, the goal of optimization for large-scale expensive global optimization (LSEGO) problems is to find satisfactory solutions within a limited number of expensive objective function evaluations. 

Coordinate descent (CD)~\cite{Coordinate_Strategy} is one of the multidimensional optimization algorithms which is called Coordinate strategy in evolutionary computation community~\cite{Coordinate_Strategy}. CD is based on the decomposition of a $n$-dimensional problem into $n$ one-dimensional sub-problems. In CD, each variable is updated as a one-dimensional optimization problem, while other variables remain fixed. CD can be seen as a special case of Block Coordinate Descent (BCD), which divides the decision space into $N$ blocks~\cite{Adaptive_coordinate}. Cooperative Co-evolution is a generalized framework of the CD and BCD, which retain a set of candidates for each dimension or block instead of a single variable~\cite{Coordinate_CC}.      

In this paper, we propose a simple but powerful CD algorithm (MCD) to solve LSEGO problems with a limited computational budget. In each step, MCD algorithm finds the region of interest and then shrinks the search space by a folding operator. Finding the region of interest is a dimension-wise operator by halving each dimension, while the folding step decreases the search space based on the region of interest found. In addition, MCD algorithm benefits from a re-order approach to change the order of dimensions to boost the algorithm.

The experiments show that MCD algorithm is able to solve LSEGO problems with a low computational budget much efficient than the state-of-the-art cooperative co-evolution algorithm. Also, it presented a satisfactory result in low dimensions with a low number of function evaluations. 

The remainder of the paper is organized as follows. Section~\ref{sec:general} explains MCD algorithm. Section~\ref{sec:exp} assesses the proposed algorithm, while Section~\ref{sec:conc} concludes the paper.

\section{The proposed algorithm}
\label{sec:general}
One of the main problems of the metaheuristic algorithms to tackle LSEGO problems is that they need evaluation of numerous candidate solutions prior to satisfying stopping conditions. If the evaluation of objectives is computationally expensive, such approaches cannot be employed in the basic form. This is commonly faced in real-world problems such as deep neural networks and computational fluid dynamics; on the other hand, there is no applicable surrogate approaches to be utilized for large-scale expensive problems. To deal with such problems, the goal is to tackle LSEGO problems with a small number of function evaluations.

In this paper, we propose a simple but efficient algorithm to solve LSEGO problems with a low computational budget. Our algorithm has two leading steps, including finding the region of interest and folding. The region of interest is responsible for finding a promising region, while the folding step shrinks the search space. The proposed algorithm in the form of pseudo-code can be seen in Algorithm~\ref{Alg1:proposed}, while more details are explained below.

\subsection{Initialization}
Similar to other metaheuristic algorithms, initialization is the first step of the MCD algorithm. To this end, one candidate solution with dimension $D$, $X=\{x_{1},x_{2},...,x_{D}\}$, is generated which their values are the center of the search space for each dimension. Besides, MCD algorithm benefits from another candidate solution, which is exactly the same as the first candidate solution in the initialization step. Another operation in the initialization step for MCD algorithm is to generate a permutation of the dimensions. To clarify, assume that we have a problem with 4 decision variables. Permutation of the dimensions generates a vector with the different number of orders such as $P=(3,2,4,1)$, which the value of the first dimension is 3, mentioning to the third dimension, $x_{3}$, in the candidate solution. 

\subsection{Finding the region of interest}
The first specific operator of the MCD algorithm is to find the region of interest. Assume that the search space for each dimension is $[L_{i},U_{i}]$ which $i$ indicates the dimension. To select each dimension, permutation vector should be used. For example, in the first step, MCD algorithm selects $i$-th element of the candidate solution where $i$ is the first value of permutation vector $P$. In each iteration to finding the region of interest, the search space for the $i$-th dimension is divided into two equal parts. In other words, the search spaces for each region are $[L_{i},(L_{i}+U_{i})/2)$ and $[(L_{i}+U_{i})/2,U_{i}]$. Then, a representative for each region should be selected. In this paper, we suggest the center of each region because it is the closest point to any point in the corresponding region that is inspired by the concept of center-based sampling strategy~\cite{center_sampling_01,center_sampling_02,DE_Deceptive_ICCSE2019,center_sampling_03,center_sampling_CEC2020,center_sampling_PSO_SMC2020}. Rahnamayan and Yang~\cite{center_sampling_01} investigated the likelihood of closeness to a random solution for a center point is much greater than the random points; in particular, for a large-scale problem, it approaches 1. As a result, center point is a good candidate to select as a representative for each region. In other words, the representatives for each region (center points) are $L_{i}+\frac{U_{i}-L_{i}}{4}$ and $U_{i}-\frac{U_{i}-L_{i}}{4}$. 

The computed center points are set to the $i$-th dimension of two candidate solutions. Afterwards, the objective function is calculated for both candidate solutions. The better candidate solution determines the region of interest for the $i$-th dimension.

\begin{figure}
	\centering
	\includegraphics[width=7.5cm]{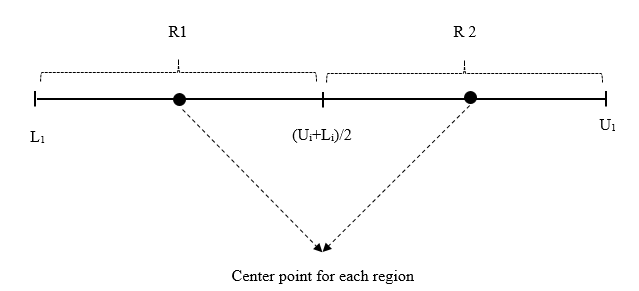}
	\caption{ The visual illustration (in 1-D) of dividing a search space into two equal regions and the corresponding center points for each region.} 
	\label{fig:center_1D}
\end{figure}

Figure.~\ref{fig:center_1D} illustrates finding the region of interest in one dimensional space. In Figure.~\ref{fig:center_1D}, $L_{1}$ and $U_{1}$ are the boundaries. As can be seen, the search space is divided into two equal regions, $R1$ and $R2$. The task of this step is to determine either $R1$ or $R2$ as the region of interest.


\subsection{Folding}
After finding the region of interest, the search space should shrink. To this end, the new upper and lower bounds are set to the upper and lower of the region of interest. Two mentioned steps (finding the region of interest and shrinkage) should be done for each dimension. 

In order to gain a better understanding, in the following, we illustrate the working of our algorithm for a simplified example in a problem with 4 decision variables, $(x_{1},x_{2},x_{3},x_{4})$ and without any permutation. Assume that search space for all variables is [-100,+100], and we solve a minimization problem. 

In the initialization phase, two same center-based candidate solutions are generated as $X=(0,0,0, 0)$ and $Y=(0,0,0,0)$.
In the next step, the region of interest for $x_{1}$ should be calculated. First, the search space for this dimension is divided into two equal sub-region including,   $[L_{1}=-100,\frac{U_{1}+L_{1}}{2}=\frac{-100+100}{2}=0]$ and  $[\frac{U_{1}+L_{1}}{2}=\frac{-100+100}{2}=0, U_{1}=+100]$. Thereby, the center point for each sub-region is -50 and +50. In the next step, two candidate solutions are changed as $X_{1}=(-50,0,0,0)$ and $X_{2}=(+50,0,0,0)$, where the values for the first dimension are the center points (-50 and +50). Assume that objective function values for $X$ and $Y$ are equal to 10 and 20, respectively. As a result, the winner candidate solution is $X$. The new search space for the first dimension is [-100,0], while for other dimensions, the search space is not changed yet. In the next step, such a process should be repeated for the second dimension, $x_{2}$. Two generated candidate solutions are $X=(-50,-50,0, 0)$ and $Y=(-50,+50,0,0)$. The value of the first dimension, $x_{1}$, is computed in the previous step, while $-50$ and $+50$ in the second dimensions are the center points for each region in the second dimension. Assume that objective function for $X_{2}$ is better than $X_{1}$. As a result, the new search space for the second dimension is $[0,100]$. Such a mechanism should be done for other remaining dimensions. This process should be repeated iteratively.

\begin{figure*}[t!]
	\centering
	\begin{subfigure}[b]{3.5cm}
		\centering
		\includegraphics[width=3.5cm] {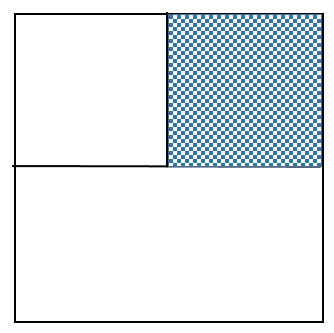}
		\caption{The first iteration}
	\end{subfigure}	
	\begin{subfigure}[b]{3.5cm}
		\centering
		\includegraphics[width=3.5cm] {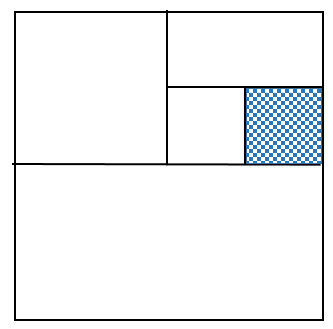}
		\caption{The second iteration}
	\end{subfigure}
	\begin{subfigure}[b]{3.4cm}
		\centering
		\includegraphics[width=3.4cm] {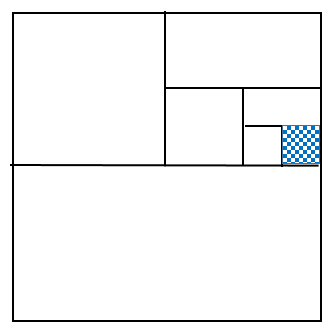}
		\caption{The third iteration}
	\end{subfigure}	
	\begin{subfigure}[b]{3.5cm}
		\centering
		\includegraphics[width=3.5cm] {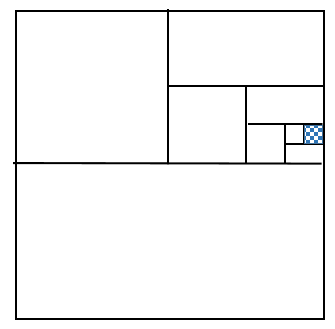}
		\caption{The 4-th iteration}
	\end{subfigure}

	\caption{The visual illustration of folding operator for a 2-D problem. In each iteration, the search space becomes a quarter ($(\frac{1}{2})^{2}$).}
	\label{fig:center_2D}
\end{figure*}

It is worthwhile to mention that in each iteration, the search space shrinks $\frac{1}{2^{D}}$ times where $D$ is the number of dimensions, and consequently, shrinking the search space after $R$ iterations is $(\frac{1}{2^{D}})^{R}$. For example, it is clear from Figure~\ref{fig:center_2D} that the region of interest is $\frac{1}{4}$ of the whole search space; or in a problem with 1000 dimensions, after only one iteration, the search space is reduced $\frac{1}{2^{1000}}$ times where is dramatically smaller than the whole search space. In other word, MCD algorithm has superior ability in the reduction of search space in each iteration exponentially. Also, another advantage of the MCD algorithm is that it is free of parameters which makes it free of control parameter tuning. 

In the MCD algorithm, we allow the algorithm to re-run, meaning that MCD algorithm can run several times with different orders (yielded by different permutation vectors) to support a high exploration capacity of the algorithm. The number of runs is a function of number of iterations for MCD algorithm and maximum number of function evaluations which is obtained as
\begin{equation}
\label{Eq:rmax}
R_{max}=max_{NFE}/(2 \times D \times max_{iter}),
\end{equation}
where $R_{max}$ is the maximum number of runs with different orders, $max_{NFE}$ is the maximum number of function evaluations, and $max_{iter}$ is the number of iterations for MCD algorithm.

\begin{algorithm2e}
	\small
	\SetAlgoLined
	\SetKwInOut{Input}{Input}\SetKwInOut{Output}{Output}
	\Input{ $D$: Dimension of problem, $Max_{NFE}$: Maximum number of expensive function evaluations, $L$: Lower bound, $U$: Upper bound, $max_{iter}$: the number of iterations of MCD algorithm }
	\Output{ $S^{*}$: the best solution found so far }
	\BlankLine
	\tcc{inf means infinitive.}
	$S^{*}=inf$; \\ 
	\tcc{MCD algorithm is run $R_{max}$ times with different orders.}
	$R_{max}=Max_{NFE}/(2*D*max_{iter});$ \\
	\For {$R\leftarrow 1$ \KwTo $R_{max}$}
	{
		\tcc{Initilization}
		$X \leftarrow$ Generate a vector with length of $D$ based on the center of the search space in each dimension; \\
		$Y \leftarrow X$; \\
		$Perm \leftarrow$ A random permutation of dimensions; \\
		\For {$iter\leftarrow 1$ \KwTo $max_{iter}$}{
			
			{
				\tcc{Finding the region of interest}
				\For {$ ind\leftarrow 1$ \KwTo $D$}{
					$i \leftarrow Perm[ind]$;\\
					$C=\frac{L_{i}+U_{i}}{2}$; and 
					$q=\frac{U_{i}-L_{i}}{4}$; \\
					
					$X[i]=L_{i}+q$; \\
					$Y[i]=U_{i}-q$; \\
					$f1 \leftarrow f(X)$; \\
					$f2 \leftarrow f(Y)$; \\
					\tcc{Folding}
					\eIf{$f1<f2$}
					{$S \leftarrow X$; \\
						$U_{i}=C$; \\	
						
					}
					{$S \leftarrow Y$;\\
						$L_{i}=C$;
					}
					$X  \leftarrow S$; \\
					$Y  \leftarrow S$; \\
					
				}
			}
		}
		\If{$F(S)<f(S^{*})$}
		{$S^{*} \leftarrow S;$}
	}
	
	\caption{The pseudo-code of MCD algorithm.}
	\label{Alg1:proposed}
\end{algorithm2e}

\section{Experimental results}
\label{sec:exp}
This section assesses MCD algorithm with limited computational budget and in comparison with cooperative co-evolution with delta grouping~\cite{CC_delta}. To this end, CEC-2010~\cite{CEC2010}, as a large-scale benchmark function set, is selected. However this benchmark function set is computationally cheap, we assume that they are computationally expensive and only a limited number of fitness evaluations using these fitness functions is allowed. It consists of 20 LSGO functions (D=1000) dividing into four categories, including separable functions (F1-F3), partially separable functions (F4-F8) which a
small number of variables are dependent, partially separable functions (F9-F18) with multiple independent subcomponents, and fully non-separable functions (F19-F20). 
We also conducted some experiments on CEC-2017~\cite{CEC2017} to investigate the behavior of MCD algorithm in lower dimensions with limited computational budget. In the last column of the tables, $w/t/l$ means MCD wins in $w$ functions, ties in $t$ functions, and loses in $l$ functions. 

\subsection{Results on large-scale expensive optimization problems }
This section benchmarks MCD algorithm on CEC-2010 LSGO benchmark functions. For the experiments, we dedicated a small computational budget to run MCD algorithm including 10,000, 20,000, and 30,000 function evaluations. According to the number function evaluations for large-scale problems, testing different order of dimensions is not applicable. Thereby, $max_{iter}$ is selected so that the value of $R_{max}$ is equal to 1, meaning that $R_{max}$ is set to 5, 10, and 15 for 10,000, 20,000, and 30,000 function evaluations, respectively. Also, for CC algorithm, population size is set to 50, while \textit{F}, \textit{CR}, and the number of sub-components are 0.5, 0.9, and 10, respectively. 

In addition, we define the Improved Accuracy Rate (IAR) for each function investigating the relative improvement that yielded by the MCD algorithm and is formulated as

\begin{equation}
IAR=\frac{Error \; of \; CC }{Error \; of \; MCD}
\end{equation}

\begin{align*}
\text{Error of MCD (or CC)} \\ = f(x)-f(x*)
\end{align*} 

Where $f(x)$ is the obtained objective function value and $f(x*)$ is the optimal value. The IAR value greater than 1 indicates that MCD algorithm performs better than CC algorithm. The IAR values greater than 1 in the tables are boldfaced. 

From Table~\ref{tab:LSGO}, we can compare the results of MCD algorithm with CC algorithm with only 10,000 function evaluations and $max_{iter}$ is set to 5, meaning that $R_{max}$ parameter value is 1. From the table, MCD algorithm outperforms CC algorithm in 16 out of 20 functions. MCD algorithm could not dominate CC algorithm in some partially separable functions (F4, F6, and F7), which a small number of variables are dependent, but the results are close to CC algorithm in more cases. In 12 cases, IAR was more than 2, which indicates MCD was better than CC more than 2 times in 12 functions. 
It is worth mentioning that when CC algorithm outperforms MCD algorithm, the results of MCD are close to CC algorithm. On the contrary, the results are drastically improved when MCD algorithm overcomes CC algorithm. For example, for F1, MCD was 735 times better than CC; or for F20 function, MCD algorithm presented 8.02E+06 times better results than CC algorithm. 

In the next experiment, we have increased $max_{NFE}$ and $max_{iter}$ to 20,000 and 10, respectively. The results can be seen in Table~\ref{tab:LSGO}. MCD algorithm was better than CC algorithm in 16 functions, and CC outperformed MCD in only 4 functions.  Comparison the results of $max_{NFE}=10,000$ and $max_{NFE}=20,000$ reveals that by increasing the number of function evaluations, the rate of reduction of the objective function value in the MCD algorithm is highly better than CC algorithm. For example, let us focus on F1 function. In CC algorithm, the objective function for F1 has been reduced from 1.17E+11 to 4.17E+10, while in MCD algorithm, it is decreased from 5.67E+07 to 6.41E+04, which readily indicates a high convergence rate of MCD algorithm.

\begin{table*}[]
	\centering	
	\caption{Numerical results of MCD Vs. CC for different function evaluations on CEC-2010 LSGO benchmark functions (D=1000).}
	\label{tab:LSGO}
	\begin{tabular}{c|ccc|ccc|ccc}
		\hline
		Functions & \multicolumn{3}{c}{$Max_{NFE}=10,000$} & \multicolumn{3}{c}{$Max_{NFE}=20,000$} & \multicolumn{3}{c}{$Max_{NFE}=30,000$} \\ \cline{2-10} 
		& MCD & CC & IAR & MCD  & CC & IAR & MCD  & CC & IAR \\ \hline
		F1 & \textbf{5.67E+07} & 4.17E+10 & \textbf{7.36E+02} & \textbf{6.41E+04} & 2.21E+10 & \textbf{3.45E+05} & \textbf{5.30E+01} & 1.32E+10 & \textbf{2.49E+08} \\
		F2 & \textbf{3.05E+03} & 1.43E+04 & \textbf{4.69} & \textbf{2.66E+03} & 1.20E+04 & \textbf{4.51} & \textbf{2.66E+03} & 1.05E+04 & \textbf{3.95} \\
		F3 & \textbf{2.85E+00} & 2.08E+01 & \textbf{7.31} & \textbf{4.09E-02} & 2.01E+01 & \textbf{491.44} & \textbf{1.17E-03} & 1.91E+01 & \textbf{1.63E04} \\
		F4 & 1.40+14 & \textbf{2.11E+14} & 0.34 & \textbf{1.20E+14} & 1.69E+14& \textbf{1.40} & \textbf{1.20E+14} & 1.47E+14 & \textbf{1.22} \\
		F5 & 5.32E+08 & \textbf{4.54E+08} & 0.85& \textbf{4.94E+08} & 3.71E+08 & \textbf{0.75} & 4.93E+08 & \textbf{3.37E+08} & 0.68 \\
		F6 & 2.10E+07 & \textbf{1.09E+07} & 0.52 & 1.98E+07 & \textbf{8.46E+06} & 0.44 & 1.97E+07 & \textbf{7.04E+06} & 0.36 \\
		F7 & 9.26E+10 & \textbf{3.95E+10} & 0.43 & 8.73E+10 & \textbf{2.65E+10} & 0.3 & 8.72E+11 & \textbf{2.02E+10} & 0.23 \\
		F8 & \textbf{2.42E+10} & 6.76E+13 & \textbf{2.79E+03} & \textbf{3.08E+08} & 1.73E+13 & \textbf{5.61E+04} & \textbf{2.64E+08} & 4.93E+12 & \textbf{1.86E+04} \\
		F9 & \textbf{2.22E+09} & 7.50E+10 & \textbf{33.80} & \textbf{1.98E+09} & 5.06E+10 & \textbf{25.60} & \textbf{1.97E+09} & 3.61E+10 & \textbf{18.32} \\
		F10 & \textbf{6.61E+03} & 1.77E+04 & \textbf{2.68} & \textbf{5.95E+03} & 1.63E+04 & \textbf{2.74} & \textbf{5.95E+03} & 1.56E+04 & \textbf{2.62} \\
		F11 & \textbf{2.06E+02} & 2.34E+02 & \textbf{1.14} & \textbf{1.90E+02} & 2.29E+02 & \textbf{1.20} & \textbf{1.90E+02} & 2.22E+02 & \textbf{1.17} \\
		F12 & \textbf{1.29E+06} & 8.22E+06 & \textbf{6.36} & \textbf{1.22E+06} & 7.24E+06 & \textbf{5.93} & \textbf{1.22E+06} & 6.81E+06 & \textbf{5.59} \\
		F13 & \textbf{3.81E+05} & 2.58E+11 & \textbf{6.78E+05} & \textbf{1.63E+04} & 8.33E+10 & \textbf{5.11E+06} & \textbf{1.54E+04} & 3.69E+10 & \textbf{2.39E+06} \\
		F14 & \textbf{3.99E+09} & 9.80E+10 & \textbf{24.54} & \textbf{3.59E+09} & 7.16E+10 & \textbf{19.97} & \textbf{3.57E+09} & 5.74E+10 & \textbf{16.06} \\
		F15 & \textbf{1.15E+04} & 1.91E+04 & \textbf{1.66} & \textbf{1.07E+04} & 1.83E+04 & \textbf{1.71} & \textbf{1.07E+04} & 1.77E+04 & \textbf{1.65} \\
		F16 & \textbf{3.91E+02} & 4.28E+02 & \textbf{1.09} & \textbf{3.66E+02} & 4.26E+02 & \textbf{1.17} & \textbf{3.65E+02} & 4.25E+02 & \textbf{1.16} \\
		F17 & \textbf{2.69E+06} & 1.66E+07 & \textbf{6.17} & \textbf{2.55E+06} & 1.38E+07 & \textbf{5.42} & \textbf{2.54E+06} & 1.28E+07 & \textbf{5.03} \\
		F18 & \textbf{1.47E+06} & 1.28E+12 & \textbf{8.50E+05} & \textbf{1.67E+05} & 5.67E+11 & \textbf{3.40E+06} & \textbf{1.85E+05} & 2.82E+11 & \textbf{1.79E+06} \\
		F19 & 2.16E+08 & \textbf{3.42E+07} & 0.15 & 1.50E+08 & \textbf{3.20E+07} & 0.21 & 1.48E+08 & \textbf{3.02E+07} & 0.20 \\
		F20 & \textbf{1.82E+05} & 1.46E+12 & \textbf{8.02E+06} & \textbf{4.65E+03} & 6.48E+11 & \textbf{1.39E+08} & \textbf{4.09E+03} & 3.19E+11 & \textbf{7.80E+07} \\ \hline
		w/t/l & \multicolumn{2}{c}{15/0/5} &  & \multicolumn{2}{c}{17/0/3} &  & \multicolumn{2}{c}{16/0/4} &  \\ \hline
	\end{tabular}
\end{table*}

By increasing the number of function evaluations and $max_{iter}$ to 30,000 and 15, similar behavior is observed between the algorithms. From Table~\ref{tab:LSGO}, MCD algorithm decreases significantly objective function value for most cases, while CC algorithm has a slight decrease during the optimization process. MCD  algorithm outperforms CC algorithm on 15 cases. 

\begin{figure*}[t!]
	\centering
	\begin{subfigure}[b]{4cm}
		\centering
		\includegraphics[width=4cm] {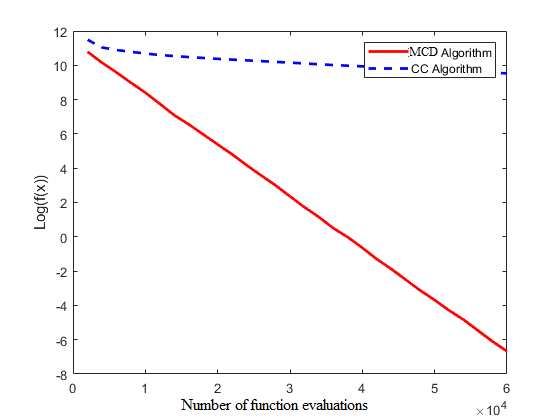}
		\caption{F1}
	\end{subfigure}	
	\begin{subfigure}[b]{4cm}
		\centering
		\includegraphics[width=4cm] {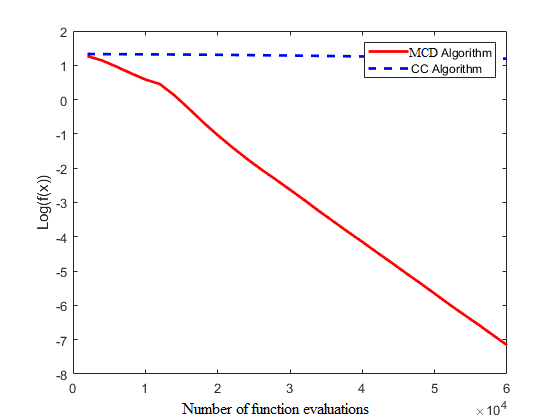}
		\caption{F3}
	\end{subfigure}
	\begin{subfigure}[b]{4cm}
		\centering
		\includegraphics[width=4cm] {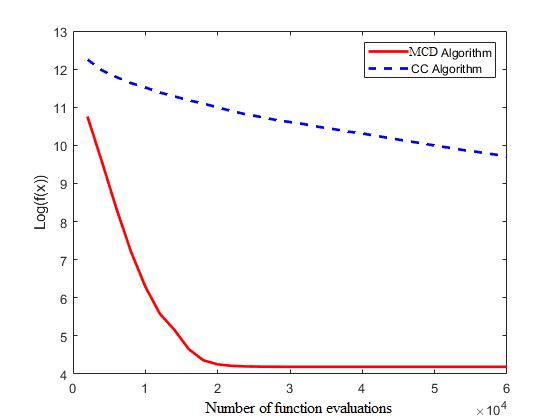}
		\caption{F13}
	\end{subfigure}	 \\
	\begin{subfigure}[b]{4cm}
		\centering
		\includegraphics[width=4cm] {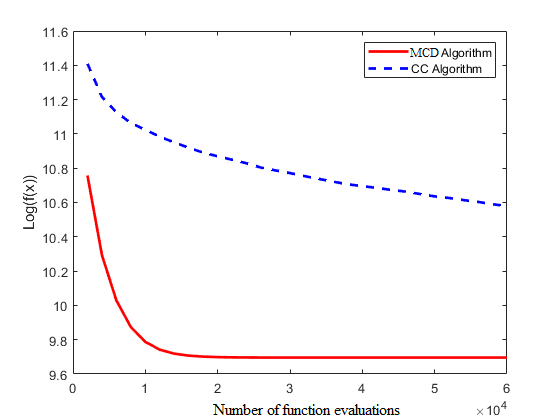}
		\caption{F14}
	\end{subfigure}	  
	\begin{subfigure}[b]{4cm}
		\centering
		\includegraphics[width=4cm] {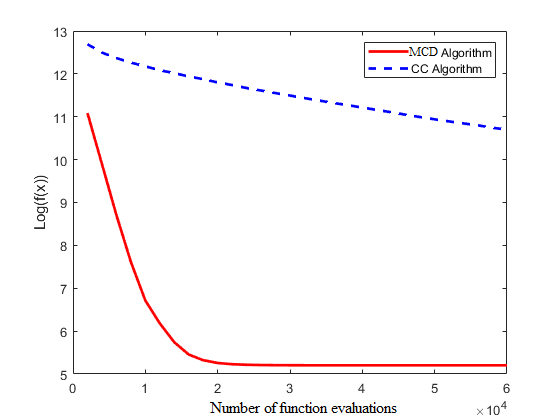}
		\caption{F18}
	\end{subfigure}
	\begin{subfigure}[b]{4cm}
		\centering
		\includegraphics[width=4cm] {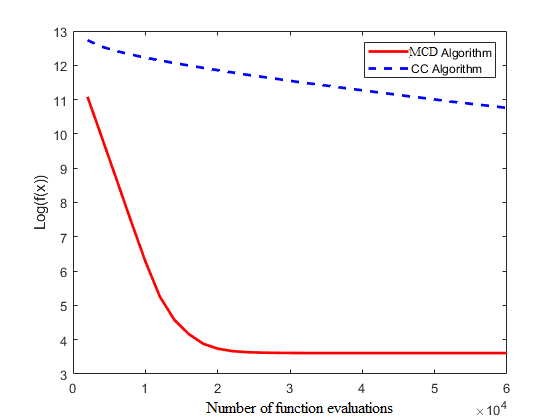}
		\caption{F20}
	\end{subfigure}	
	
	\caption{Convergence plot of some selected functions for MCD and CC algorithms, D=1000 and $Max_{NFS}=60,000$.}
	\label{fig:conv_curve}
\end{figure*}

Figure.~\ref{fig:conv_curve} indicates the convergence curve for some selected functions, which shows a high convergence rate of MCD algorithm in comparison to CC algorithm. From the figure, the convergence curve for unimodal functions (F1 and F3) are straight with a steep slope, which indicates the power of MCD algorithm to tackle unimodal functions. For multi-modal functions, initially, the slope is very sharp, and the slope gets almost flat in the later stages.
\subsection{Results on low-scale expensive optimization problems }

Last but not least, we carried out some experiments to investigate the behavior of MCD algorithm and in comparison to DE algorithm in lower dimensions. To this end, CEC-2017~\cite{CEC2017} is selected which includes 30 benchmark functions with different characteristics including unimodal functions (F1-F3), multimodal functions (F4-F10), hybrid multi-modal functions (F11-F20) and composite functions (F21-F30). For the experiments, $D$ is set to 10, 30, 50, and 100. In addition, for DE algorithm, $CR$ is set to 0.9, while $F$ is a random number between 0.2 and 0.8. In all experiments, $max_{iter}$ is set to 10. Accordingly, the number of re-runs can be calculated using Eq.~\ref{Eq:rmax}.

In the first experiment, maximum number of function evaluations is set to $100 \times D$ and consequently, $R_{max}$ is 5. The numerical results can be seen in Table~\ref{CEC2017_100D}. From the table, in all dimensions, MCD outperforms DE. For $D=10$, MCD yielded better results in 28 out of 30 functions, while for $D=30$, $50$, and $D=100$ MCD wins in 29 functions. From the table, IAR is more than 2 in most cases, showing that MCD was better than DE more than 2 times in most cases. In other words, for $D=10, 30, 50$ and $100$, MCD was better than 2 times in 17, 20, 20, and 21 cases.  
Thereby, we can say that MCD algorithm overcomes DE algorithm with a limited computation budget in lower dimensions. 

\begin{table*}[]
	\centering
	\caption{Numerical results of MCD Vs. DE for $100 \times D$ function evaluations on CEC-2017 benchmark functions.}
	\label{CEC2017_100D}
	\begin{tabular}{c|p{0.9cm}p{0.9cm}p{1cm}|p{0.9cm}p{0.9cm}p{1cm}|p{0.9cm}p{0.9cm}p{1cm}|p{0.9cm}p{0.9cm}p{0.9cm}}
		\hline
		Functions & \multicolumn{3}{c}{D=10} & \multicolumn{3}{c}{D=30} & \multicolumn{3}{c}{D=50} & \multicolumn{3}{c}{D=100} \\ \cline{2-13} 
		& MCD & DE & IAR & MCD & DE & IAR & MCD & DE & IAR & MCD & DE & IAR \\ \hline
		F1 & \textbf{1.42E+06} & 3.49E+09 & \textbf{2458.47} & \textbf{5.34E+07} & 2.74E+10 & \textbf{513.62} & \textbf{2.42E+08} & 5.98E+10 & \textbf{247.23} & \textbf{4.65E+08} & 1.37E+11 & \textbf{295.15} \\
		F2 & \textbf{3.35E+07} & 1.16E+10 & \textbf{3.45E+02} & \textbf{7.77E+29} & 2.01E+38 & \textbf{2.59E+08} & \textbf{5.85E+50} & 3.58E+72 & \textbf{6.12E+21} & \textbf{8.52E+124} & 1.82E+163 & \textbf{2.13E+38} \\
		F3 & \textbf{1.79E+04} & 2.95E+04 & \textbf{1.65} & \textbf{1.17E+05} & 2.14E+05 & \textbf{1.82} & \textbf{2.23E+05} & 4.08E+05 & \textbf{1.83} & \textbf{3.80E+05} & 8.60E+05 & \textbf{2.26} \\
		F4 & \textbf{1.58E+01} & 2.32E+02 & \textbf{14.64} & \textbf{1.64E+02} & 3.33E+03 & \textbf{20.27} & \textbf{3.47E+02} & 9.57E+03 & \textbf{27.60} & \textbf{7.00E+02} & 3.04E+03 & \textbf{4.35} \\
		F5 & \textbf{2.27E+01} & 7.92E+01 & \textbf{3.48} & \textbf{1.03E+02} & 3.53E+02 & \textbf{3.42} & \textbf{1.98E+02} & 6.54E+02 & \textbf{3.30} & \textbf{4.88E+02} & 1.08E+03 & \textbf{2.22} \\
		F6 & \textbf{4.19E+00} & 4.74E+01 & \textbf{11.30} & \textbf{5.59E+00} & 6.98E+01 & \textbf{12.49} & \textbf{5.54E+00} & 8.08E+01 & \textbf{14.58} & \textbf{6.03E+00} & 4.39E+01 & \textbf{7.28} \\
		F7 & \textbf{2.90E+01} & 2.09E+02 & \textbf{7.21} & \textbf{1.50E+02} & 1.06E+03 & \textbf{7.09} & \textbf{2.69E+02} & 2.00E+03 & \textbf{7.43} & \textbf{6.40E+02} & 1.72E+03 & \textbf{2.69} \\
		F8 & \textbf{2.10E+01} & 8.40E+01 & \textbf{3.99} & \textbf{1.02E+02} & 3.50E+02 & \textbf{3.42} & \textbf{2.14E+02} & 6.37E+02 & \textbf{2.98} & \textbf{4.87E+02} & 1.09E+03 & \textbf{2.23} \\
		F9 & \textbf{1.46E+02} & 1.52E+03 & \textbf{10.39} & \textbf{2.15E+03} & 1.24E+04 & \textbf{5.77} & \textbf{4.94E+03} & 3.11E+04 & \textbf{6.30} & \textbf{1.49E+04} & 2.04E+04 & \textbf{1.37} \\
		F10 & \textbf{1.23E+03} & 2.04E+03 & \textbf{1.66} & \textbf{3.77E+03} & 8.36E+03 & \textbf{2.22} & \textbf{6.41E+03} & 1.50E+04 & \textbf{2.33} & \textbf{1.25E+04} & 3.15E+04 & \textbf{2.53} \\
		F11 & 1.38E+03 & \textbf{7.68E+02} & 0.56 & \textbf{8.57E+03} & 1.24E+04 & \textbf{1.45} & \textbf{2.00E+04} & 3.78E+04 & \textbf{1.88} & \textbf{8.77E+04} & 2.92E+05 & \textbf{3.33} \\
		F12 & \textbf{1.51E+05} & 1.26E+08 & \textbf{837.17} & \textbf{1.10E+07} & 1.97E+09 & \textbf{178.66} & \textbf{9.83E+07} & 9.99E+09 & \textbf{101.63} & \textbf{5.72E+08} & 4.66E+09 & \textbf{8.15} \\
		F13 & \textbf{1.75E+04} & 3.96E+05 & \textbf{22.69} & \textbf{2.36E+06} & 4.72E+08 & \textbf{199.80} & \textbf{1.11E+08} & 2.13E+09 & \textbf{19.29} & \textbf{9.31E+04} & 6.77E+07 & \textbf{726.84} \\
		F14 & \textbf{2.45E+02} & 2.00E+03 & \textbf{8.18} & 2.66E+06 & \textbf{9.93E+05} & 0.37 & 1.54E+07 & \textbf{4.62E+06} & 0.30 & \textbf{1.73E+07} & 2.72E+07 & \textbf{1.57} \\
		F15 & \textbf{1.46E+03} & 1.49E+04 & \textbf{10.20} & \textbf{1.40E+05} & 7.92E+07 & \textbf{564.95} & \textbf{1.15E+07} & 2.28E+08 & \textbf{19.91} & \textbf{5.54E+04} & 6.17E+06 & \textbf{111.33} \\
		F16 & \textbf{4.10E+02} & 4.72E+02 & \textbf{1.15} & \textbf{1.10E+03} & 2.48E+03 & \textbf{2.26} & \textbf{1.67E+03} & 4.54E+03 & \textbf{2.72} & \textbf{3.53E+03} & 9.47E+03 & \textbf{2.68} \\
		F17 & 3.78E+02 & \textbf{2.14E+02} & 0.57 & \textbf{4.77E+02} & 1.11E+03 & \textbf{2.32} & \textbf{1.47E+03} & 3.15E+03 & \textbf{2.15} & \textbf{2.97E+03} & 6.28E+03 & \textbf{2.11} \\
		F18 & \textbf{3.49E+03} & 2.06E+06 & \textbf{589.68} & \textbf{5.65E+06} & 2.23E+07 & \textbf{3.95} & \textbf{2.19E+07} & 5.39E+07 & \textbf{2.46} & \textbf{1.45E+07} & 6.11E+07 & \textbf{4.21} \\
		F19 & \textbf{2.83E+03} & 4.99E+04 & \textbf{17.61} & \textbf{3.62E+04} & 8.92E+07 & \textbf{2464.22} & \textbf{2.98E+04} & 1.19E+08 & \textbf{3988.17} & \textbf{5.69E+04} & 1.70E+07 & \textbf{298.02} \\
		F20 & \textbf{1.83E+01} & 2.62E+02 & \textbf{14.29} & \textbf{6.19E+02} & 1.15E+03 & \textbf{1.86} & \textbf{1.20E+03} & 2.54E+03 & \textbf{2.11} & \textbf{2.53E+03} & 5.82E+03 & \textbf{2.30} \\
		F21 & \textbf{2.23E+02} & 2.73E+02 & \textbf{1.22} & \textbf{3.17E+02} & 5.35E+02 & \textbf{1.69} & \textbf{4.01E+02} & 8.30E+02 & \textbf{2.07} & \textbf{7.06E+02} & 1.30E+03 & \textbf{1.84} \\
		F22 & \textbf{1.17E+02} & 5.02E+02 & \textbf{4.31} & \textbf{1.40E+02} & 7.61E+03 & \textbf{54.39} & \textbf{7.39E+02} & 1.53E+04 & \textbf{20.66} & \textbf{1.39E+03} & 3.26E+04 & \textbf{23.39} \\
		F23 & \textbf{3.34E+02} & 3.78E+02 & \textbf{1.13} & \textbf{4.90E+02} & 6.98E+02 & \textbf{1.42} & \textbf{7.30E+02} & 1.11E+03 & \textbf{1.52} & \textbf{9.47E+02} & 1.61E+03 & \textbf{1.70} \\
		F24 & \textbf{2.43E+02} & 4.12E+02 & \textbf{1.69} & \textbf{5.29E+02} & 7.45E+02 & \textbf{1.41} & \textbf{7.54E+02} & 1.11E+03 & \textbf{1.47} & \textbf{1.30E+03} & 1.98E+03 & \textbf{1.53} \\
		F25 & \textbf{4.38E+02} & 6.46E+02 & \textbf{1.48} & \textbf{5.12E+02} & 2.37E+03 & \textbf{4.63} & \textbf{8.05E+02} & 8.40E+03 & \textbf{10.44} & \textbf{1.58E+03} & 4.39E+03 & \textbf{2.77} \\
		F26 & \textbf{4.56E+02} & 8.97E+02 & \textbf{1.97} & \textbf{1.48E+03} & 5.08E+03 & \textbf{3.44} & \textbf{2.86E+03} & 8.56E+03 & \textbf{2.99} & \textbf{8.78E+03} & 1.45E+04 & \textbf{1.65} \\
		F27 & \textbf{4.11E+02} & 4.25E+02 & \textbf{1.03} & \textbf{5.54E+02} & 7.22E+02 & \textbf{1.30} & \textbf{8.60E+02} & 1.40E+03 & \textbf{1.62} & \textbf{8.68E+02} & 1.35E+03 & \textbf{1.55} \\
		F28 & \textbf{5.87E+02} & 6.49E+02 & \textbf{1.11} & \textbf{6.58E+02} & 2.49E+03 & \textbf{3.79} & \textbf{1.61E+03} & 7.10E+03 & \textbf{4.41} & \textbf{2.60E+03} & 9.14E+03 & \textbf{3.51} \\
		F29 & \textbf{4.44E+02} & 4.85E+02 & \textbf{1.09} & \textbf{1.31E+03} & 2.51E+03 & \textbf{1.92} & \textbf{2.64E+03} & 5.08E+03 & \textbf{1.92} & \textbf{5.37E+03} & 8.05E+03 & \textbf{1.50} \\
		F30 & \textbf{1.57E+06} & 6.30E+06 & \textbf{4.02} & \textbf{1.03E+07} & 6.94E+07 & \textbf{6.72} & \textbf{5.18E+08} & 6.32E+08 & \textbf{1.22} & 5.23E+07 & \textbf{4.66E+07} & 0.89 \\ \hline
		w/t/l & \multicolumn{2}{c}{28/0/2} &  & \multicolumn{2}{c}{29/0/1} &  & \multicolumn{2}{c}{29/0/1} &  & \multicolumn{2}{c}{29/0/1} &  \\ \hline
	\end{tabular}
\end{table*}

In the next experiment, we increase the number of function evaluations to $500 \times D$ to investigate the behavior of MCD algorithm with a more computation budget. From Table~\ref{CEC_2017_500D}, MCD algorithm has retained its performance in most cases. MCD algorithm was better than DE in 27 functions for $D=30$ and 25 functions for $D=50$, while for $D=100$, MCD outperformed DE in 28 cases.

\begin{table*}[]
	\centering
	\caption{Numerical results of MCD Vs. DE for $500 \times D$ function evaluations on CEC-2017 benchmark functions.}
	\label{CEC_2017_500D}
	\begin{tabular}{c|p{0.9cm}p{0.9cm}p{01cm}|p{0.9cm}p{0.9cm}p{1cm}|p{0.9cm}p{0.9cm}p{1cm}|p{0.9cm}p{0.9cm}p{1cm}}
		\hline
		\multirow{2}{*}{Functions} & \multicolumn{3}{c}{D=10} & \multicolumn{3}{c}{D=30} & \multicolumn{3}{c}{D=50} & \multicolumn{3}{c}{D=100} \\ \cline{2-13} 
		& MCD & DE & IAR & MCD & DE & IAR & MCD & DE & IAR & MCD & DE & IAR \\ \hline
		F1 & \textbf{9.42E+04} & 8.53E+07 & \textbf{905.63} & \textbf{2.02E+07} & 1.26E+09 & \textbf{62.31} & \textbf{1.03E+08} & 3.50E+09 & \textbf{33.80} & \textbf{1.84E+08} & 6.07E+09 & \textbf{33.03} \\
		F2 & \textbf{1.08E+06} & 7.13E+06 & \textbf{6.59E+00} & \textbf{8.54E+26} & 1.25E+32 & \textbf{1.46E+05} & \textbf{4.32E+47} & 7.73E+63 & \textbf{1.79E+16} & \textbf{1.39E+118} & 4.73E+148 & \textbf{3.39E+30} \\
		F3 & \textbf{1.29E+04} & 1.84E+04 & \textbf{1.42} & \textbf{8.96E+04} & 1.81E+05 & \textbf{2.02} & \textbf{1.82E+05} & 3.49E+05 & \textbf{1.91} & \textbf{3.68E+05} & 8.69E+05 & \textbf{2.36} \\
		F4 & \textbf{6.15E+00} & 1.34E+01 & \textbf{2.18} & \textbf{1.39E+02} & 2.38E+02 & \textbf{1.71} & \textbf{2.84E+02} & 6.18E+02 & \textbf{2.17} & \textbf{6.65E+02} & 1.07E+03 & \textbf{1.61} \\
		F5 & \textbf{1.68E+01} & 4.89E+01 & \textbf{2.91} & \textbf{7.75E+01} & 2.43E+02 & \textbf{3.14} & \textbf{1.70E+02} & 4.63E+02 & \textbf{2.72} & \textbf{4.78E+02} & 1.01E+03 & \textbf{2.11} \\
		F6 & \textbf{2.40E+00} & 1.04E+01 & \textbf{4.35} & \textbf{3.88E+00} & 2.21E+01 & \textbf{5.69} & \textbf{4.42E+00} & 2.57E+01 & \textbf{5.80} & \textbf{5.90E+00} & 2.34E+01 & \textbf{3.97} \\
		F7 & \textbf{2.08E+01} & 7.23E+01 & \textbf{3.47} & \textbf{1.22E+02} & 3.43E+02 & \textbf{2.81} & \textbf{2.44E+02} & 6.11E+02 & \textbf{2.51} & \textbf{6.04E+02} & 1.28E+03 & \textbf{2.12} \\
		F8 & \textbf{1.58E+01} & 5.03E+01 & \textbf{3.19} & \textbf{8.50E+01} & 2.54E+02 & \textbf{2.99} & \textbf{1.79E+02} & 4.70E+02 & \textbf{2.63} & \textbf{4.62E+02} & 1.02E+03 & \textbf{2.20} \\
		F9 & \textbf{7.88E+01} & 1.22E+02 & \textbf{1.55} & \textbf{1.15E+03} & 2.02E+03 & \textbf{1.75} & \textbf{3.00E+03} & 3.83E+03 & \textbf{1.27} & 1.44E+04 & 8.06E+03 & 0.56 \\
		F10 & \textbf{9.52E+02} & 1.63E+03 & \textbf{1.71} & \textbf{3.51E+03} & 7.84E+03 & \textbf{2.23} & \textbf{5.59E+03} & 1.44E+04 & \textbf{2.58} & \textbf{1.22E+04} & 3.14E+04 & \textbf{2.56} \\
		F11 & 6.47E+02 & \textbf{5.11E+01} & 0.08 & 4.45E+03 & \textbf{1.18E+03} & 0.27 & 1.23E+04 & 6.03E+03 & 0.49 & \textbf{8.90E+04} & 2.40E+05 & \textbf{2.69} \\
		F12 & \textbf{2.20E+04} & 8.36E+06 & \textbf{380.04} & \textbf{5.00E+06} & 1.48E+08 & \textbf{29.63} & \textbf{3.66E+07} & 7.35E+08 & \textbf{20.09} & \textbf{5.72E+08} & 1.46E+09 & \textbf{2.55} \\
		F13 & 1.71E+04 & \textbf{6.43E+03} & 0.38 & \textbf{1.34E+05} & 4.46E+06 & \textbf{33.33} & 1.64E+07 & 1.55E+07 & 0.94 & \textbf{8.59E+04} & 4.09E+06 & \textbf{47.66} \\
		F14 & 2.42E+02 & \textbf{2.16E+02} & 0.90 & 7.07E+05 & \textbf{1.02E+05} & 0.14 & 3.02E+06 & 9.52E+05 & 0.31 & \textbf{1.38E+07} & 1.84E+07 & \textbf{1.33} \\
		F15 & \textbf{1.35E+03} & 1.58E+03 & \textbf{1.17} & \textbf{2.36E+04} & 9.13E+05 & \textbf{38.63} & \textbf{6.31E+05} & 8.76E+05 & \textbf{1.39} & \textbf{5.17E+04} & 1.67E+05 & \textbf{3.24} \\
		F16 & \textbf{1.55E+02} & 1.97E+02 & \textbf{1.27} & \textbf{7.73E+02} & 1.86E+03 & \textbf{2.41} & \textbf{1.40E+03} & 3.77E+03 & \textbf{2.69} & \textbf{3.36E+03} & 9.16E+03 & \textbf{2.72} \\
		F17 & 3.54E+02 & \textbf{1.04E+02} & 0.29 & \textbf{2.64E+02} & 8.77E+02 & \textbf{3.32} & \textbf{1.07E+03} & 2.46E+03 & \textbf{2.31} & \textbf{2.92E+03} & 6.04E+03 & \textbf{2.07} \\
		F18 & \textbf{2.28E+03} & 4.04E+04 & \textbf{17.67} & \textbf{7.41E+05} & 4.77E+06 & \textbf{6.43} & \textbf{7.29E+06} & 1.21E+07 & \textbf{1.66} & \textbf{1.11E+07} & 4.71E+07 & \textbf{4.24} \\
		F19 & 2.83E+03 & \textbf{1.10E+03} & 0.39 & \textbf{2.21E+04} & 1.72E+06 & \textbf{77.84} & \textbf{1.77E+04} & 6.26E+05 & \textbf{35.45} & \textbf{5.53E+04} & 3.02E+05 & \textbf{5.46} \\
		F20 & \textbf{1.54E+01} & 1.13E+02 & \textbf{7.33} & \textbf{4.02E+02} & 9.08E+02 & \textbf{2.26} & \textbf{9.07E+02} & 2.16E+03 & \textbf{2.38} & \textbf{2.39E+03} & 5.73E+03 & \textbf{2.40} \\
		F21 & \textbf{2.03E+02} & 2.30E+02 & \textbf{1.13} & \textbf{2.91E+02} & 4.44E+02 & \textbf{1.53} & \textbf{3.81E+02} & 6.64E+02 & \textbf{1.74} & \textbf{6.91E+02} & 1.24E+03 & \textbf{1.80} \\
		F22 & \textbf{1.06E+02} & 1.19E+02 & \textbf{1.12} & \textbf{1.32E+02} & 4.21E+03 & \textbf{31.94} & \textbf{2.62E+02} & 1.44E+04 & \textbf{54.87} & \textbf{9.47E+02} & 3.23E+04 & \textbf{34.14} \\
		F23 & \textbf{3.15E+02} & 3.44E+02 & \textbf{1.09} & \textbf{4.33E+02} & 5.87E+02 & \textbf{1.35} & \textbf{6.73E+02} & 8.93E+02 & \textbf{1.33} & \textbf{9.40E+02} & 1.55E+03 & \textbf{1.64} \\
		F24 & \textbf{1.75E+02} & 3.75E+02 & \textbf{2.14} & \textbf{4.53E+02} & 6.52E+02 & \textbf{1.44} & \textbf{7.06E+02} & 9.42E+02 & \textbf{1.33} & \textbf{1.27E+03} & 1.90E+03 & \textbf{1.50} \\
		F25 & \textbf{4.11E+02} & 4.49E+02 & \textbf{1.09} & \textbf{4.52E+02} & 5.19E+02 & \textbf{1.15} & \textbf{7.01E+02} & 9.53E+02 & \textbf{1.36} & \textbf{1.50E+03} & 2.01E+03 & \textbf{1.34} \\
		F26 & \textbf{2.87E+02} & 3.89E+02 & \textbf{1.36} & \textbf{8.29E+02} & 3.55E+03 & \textbf{4.29} & \textbf{1.86E+03} & 5.79E+03 & \textbf{3.11} & \textbf{7.17E+03} & 1.35E+04 & \textbf{1.89} \\
		F27 & 4.02E+02 & \textbf{3.96E+02} & 0.99 & \textbf{5.41E+02} & 5.51E+02 & \textbf{1.02} & 7.90E+02 & \textbf{7.70E+02} & 0.97 & \textbf{8.55E+02} & 1.13E+03 & \textbf{1.32} \\
		F28 & \textbf{4.43E+02} & 4.59E+02 & \textbf{1.04} & \textbf{5.84E+02} & 6.15E+02 & \textbf{1.05} & 1.27E+03 & 1.60E+03 & \textbf{1.26} & \textbf{2.34E+03} & 4.60E+03 & \textbf{1.97} \\
		F29 & \textbf{3.46E+02} & 3.61E+02 & \textbf{1.04} & \textbf{9.85E+02} & 1.56E+03 & \textbf{1.58} & 2.21E+03 & 3.04E+03 & \textbf{1.37} & \textbf{5.26E+03} & 7.34E+03 & \textbf{1.40} \\
		F30 & \textbf{3.53E+05} & 6.96E+05 & \textbf{1.97} & 2.37E+06 & \textbf{1.91E+06} & 0.81 & 3.20E+08 & 7.65E+07 & 0.24 & 4.36E+07 & \textbf{8.68E+06} & 0.20 \\ \hline
		w/t/l & \multicolumn{2}{c}{24/0/6} &  & \multicolumn{2}{c}{27/0/3} & & \multicolumn{2}{c}{25/0/5} &  & \multicolumn{2}{c}{28/0/2} &  \\ \hline
	\end{tabular}
\end{table*}

\begin{table*}[]
	\centering
	\caption{IAR values greater than 1 and 2 for different dimensions ($D$) and different number of maximum evaluations ($Max_{NFE}$).}
	\label{tab:summerization}
	\begin{tabular}{c|cc|cc|cc|cc}
		\hline
		\multirow{2}{*}{$Max_{NFE}$} & \multicolumn{2}{c}{D=10} & \multicolumn{2}{c}{D=30} & \multicolumn{2}{c}{D=50} & \multicolumn{2}{c}{D=100} \\ \cline{2-9} 
		& IAR\textgreater{}1 & IAR\textgreater{}2 & IAR\textgreater{}1 & IAR\textgreater{}2 & IAR\textgreater{}1 & IAR\textgreater{}2 & IAR\textgreater{}1 & IAR\textgreater{}2 \\ \hline
		$100 \times D$ & 28 & 12 & 29 & 19 & 29 & 20 & 29 & 21 \\
		$500 \times D$ & 24 & 11 & 27 & 18 & 25 & 15 & 28 & 19 \\ \hline
	\end{tabular}
\end{table*}

Tables~\ref{tab:summerization} shows a top-level view from the results on CEC-2017 benchmark functions. It compares the IAR values greater than 1 (the number of winner) and 2 for all dimensions and maximum number of function evaluations. From Table~\ref{tab:summerization} as well as Tables~\ref{CEC2017_100D} and~\ref{CEC_2017_500D} , we can observe the power of MCD algorithm on expensive optimization problems with low computation budget. Two main conclusions can be observed from the results are as follows:
\begin{itemize}
	\item by increasing the number of dimensions, the efficiency of MCD increases comparing to DE. It verifies that MCD is more beneficial in large-scale optimization problems so that increasing the size of the search space does not any negative effect on the performance of MCD algorithm, similar to other metaheuristic methods.
	\item by decreasing the number of fitness functions in the experiments, MCD reaches a better solution comparing to DE. That is an indicative that MCD is more proper for highly-cost optimization problems which there is not possible to consider numerous fitness evaluations for them. 
\end{itemize}

All in all, the extensive set of experiments on both CEC-2010 and CEC-2017 benchmark functions confirm that MCD algorithm is a powerful algorithm in solving expensive optimization problems in particular large-scale ones with a limited computation budget.

\section{Conclusion remarks} 
\label{sec:conc}
Metaheuristic algorithms are so prevalent in solving large-scale global optimization (LSGO) problems, while they need a large number of function evaluations. As a result, they are not affordable to employ in real-world applications. In this paper, we have proposed a modified coordinate descent algorithm (MCD) solve large-scale global optimization (LSGO) problems with a low computational budget. MCD algorithm finds a region of interest in the whole search space. Then, the search space shrinks based on the region of interest detected in the previous step. One of the main characteristics of the MCD algorithm is that it is free of parameters, which makes it free of tuning compared with other metaheuristic algorithms with minimum three control parameters.

MCD algorithm is compared on CEC-2010 LSGO benchmark functions in comparison to a cooperative co-evolution (CC) algorithm with delta grouping. Also, to investigate the behavior of MCD algorithm in lower dimensions, we carried out some experiments on CEC-2017 benchmark functions and in comparison to DE algorithm. For all experiments, we employed a limited number of function evaluations. The results indicate the competence of MCD algorithm in solving LSEGO problems with limited computational budgets. Also, comparing the results in different dimensions clearly investigates that MCD is more beneficial in higher dimensions. In other words, increasing the number of dimensions has not any adverse effects on MCD algorithm, unlike typical metaheuristic algorithms.

The authors believe that MCD algorithm has a great capability in solving LSEGO problems, while it has not been seen properly so far in the community. In the future, the authors intend to extend this work on real-world applications such as finding optimal parameters in deep networks. Also, combining MCD algorithm with other LSGO algorithms is another research direction. Proposing large-scale multi-objective extension of MCD algorithm are under investigation as well.

\bibliographystyle{IEEEbib}
\bibliography{jalal}

\begin{thebibliography}{10}

\bibitem{LSGO_benchmark}
Mirjam~Sepesy Mau{\v{c}}ec and Janez Brest,
\newblock ``A review of the recent use of differential evolution for
  large-scale global optimization: an analysis of selected algorithms on the
  {CEC} 2013 lsgo benchmark suite,''
\newblock {\em Swarm and Evolutionary Computation}, vol. 50, pp. 100428, 2019.

\bibitem{LSGO_scheduling_01}
Mehmet Erdem and Serol Bulkan,
\newblock ``A two-stage solution approach for the large-scale home healthcare
  routeing and scheduling problem,''
\newblock {\em South African Journal of Industrial Engineering}, vol. 28, no.
  4, pp. 133--149, 2017.

\bibitem{LSGO_scheduling_02}
Jo{\~a}o Soares, Mohammad Ali~Fotouhi Ghazvini, Marco Silva, and Zita Vale,
\newblock ``Multi-dimensional signaling method for population-based
  metaheuristics: Solving the large-scale scheduling problem in smart grids,''
\newblock {\em Swarm and Evolutionary Computation}, vol. 29, pp. 13--32, 2016.

\bibitem{Evolutionary_Deep}
Nasser~R Sabar, Ayad Turky, Andy Song, and Abdul Sattar,
\newblock ``An evolutionary hyper-heuristic to optimise deep belief networks
  for image reconstruction,''
\newblock {\em Applied Soft Computing}, p. 105510, 2019.

\bibitem{LSGO_vehicle_01}
Arie Chandra,
\newblock ``Optimization of very large scale capacitated vehicle routing
  problems,''
\newblock in {\em 5th International Conference on Industrial and Business
  Engineering}, 2019, pp. 18--22.

\bibitem{SA_Main_Paper}
Stephen~P Brooks and Byron~JT Morgan,
\newblock ``Optimization using simulated annealing,''
\newblock {\em Journal of the Royal Statistical Society: Series D (The
  Statistician)}, vol. 44, no. 2, pp. 241--257, 1995.

\bibitem{PSO_Main_Paper}
Yuhui Shi and Russell Eberhart,
\newblock ``A modified particle swarm optimizer,''
\newblock in {\em IEEE International Conference on Evolutionary Computation},
  1998, pp. 69--73.

\bibitem{PSO_Main_Paper02}
James Kennedy and Russell Eberhart,
\newblock ``Particle swarm optimization ({PSO}),''
\newblock in {\em IEEE International Conference on Neural Networks}, 1995, pp.
  1942--1948.

\bibitem{DE_Original}
Rainer Storn and Kenneth Price,
\newblock ``Differential evolution--a simple and efficient heuristic for global
  optimization over continuous spaces,''
\newblock {\em Journal of Global Optimization}, vol. 11, no. 4, pp. 341--359,
  1997.

\bibitem{HMS_Main_Paper}
Seyed~Jalaleddin Mousavirad and Hossein Ebrahimpour-Komleh,
\newblock ``Human mental search: a new population-based metaheuristic
  optimization algorithm,''
\newblock {\em Applied Intelligence}, vol. 47, no. 3, pp. 850--887, 2017.

\bibitem{GHMS-RCS}
Seyed~Jalaleddin Mousavirad, Gerald Schaefer, and Iakov Korovin,
\newblock ``A global-best guided human mental search algorithm with random
  clustering strategy,''
\newblock in {\em International Conference on Systems, Man and Cybernetics}.
  IEEE, 2019, pp. 3174--3179.

\bibitem{large_scale_survey}
Sedigheh Mahdavi, Mohammad~Ebrahim Shiri, and Shahryar Rahnamayan,
\newblock ``Metaheuristics in large-scale global continues optimization: A
  survey,''
\newblock {\em Information Sciences}, vol. 295, pp. 407--428, 2015.

\bibitem{LSGO_ICC}
Sedigheh Mahdavi, Shahryar Rahnamayan, and Mohammad~Ebrahim Shiri,
\newblock ``Incremental cooperative coevolution for large-scale global
  optimization,''
\newblock {\em Soft Computing}, vol. 22, no. 6, pp. 2045--2064, 2018.

\bibitem{LSGO_CCS}
Weiming Liu, Yinda Zhou, Bin Li, and Ke~Tang,
\newblock ``Cooperative co-evolution with soft grouping for large scale global
  optimization,''
\newblock in {\em IEEE Congress on Evolutionary Computation}. IEEE, 2019, pp.
  318--325.

\bibitem{LSGO_CC01}
Zhenyu Yang, Ke~Tang, and Xin Yao,
\newblock ``Large scale evolutionary optimization using cooperative
  coevolution,''
\newblock {\em Information sciences}, vol. 178, no. 15, pp. 2985--2999, 2008.

\bibitem{CC_survey}
Xiaoliang Ma, Xiaodong Li, Qingfu Zhang, Ke~Tang, Zhengping Liang, Weixin Xie,
  and Zexuan Zhu,
\newblock ``A survey on cooperative co-evolutionary algorithms,''
\newblock {\em IEEE Transactions on Evolutionary Computation}, vol. 23, no. 3,
  pp. 421--441, 2018.

\bibitem{CC_main_paper01}
Mitchell~A Potter,
\newblock {\em The design and analysis of a computational model of cooperative
  coevolution},
\newblock Ph.D. thesis, Citeseer, 1997.

\bibitem{CC_main_paper02}
Mitchell~A Potter and Kenneth~A De~Jong,
\newblock ``A cooperative coevolutionary approach to function optimization,''
\newblock in {\em International Conference on Parallel Problem Solving from
  Nature}. Springer, 1994, pp. 249--257.

\bibitem{center_sampling_02}
Hanan Hiba, Sedigheh Mahdavi, and Shahryar Rahnamayan,
\newblock ``Differential evolution with center-based mutation for large-scale
  optimization,''
\newblock in {\em IEEE Symposium Series on Computational Intelligence}. IEEE,
  2017, pp. 1--8.

\bibitem{center_sampling_05}
Hanan Hiba, Amin Ibrahim, and Shahryar Rahnamayan,
\newblock ``Large-scale optimization using center-based differential evolution
  with dynamic mutation scheme,''
\newblock in {\em IEEE Congress on Evolutionary Computation}. IEEE, 2019, pp.
  3189--3196.

\bibitem{LSGO_PSO01}
Qiang Yang, Wei-Neng Chen, Tianlong Gu, Huaxiang Zhang, Huaqiang Yuan, Sam
  Kwong, and Jun Zhang,
\newblock ``A distributed swarm optimizer with adaptive communication for
  large-scale optimization,''
\newblock {\em IEEE transactions on cybernetics}, 2019.

\bibitem{LSGO_Expensive_01}
Yaochu Jin and Bernhard Sendhoff,
\newblock ``A systems approach to evolutionary multiobjective structural
  optimization and beyond,''
\newblock {\em IEEE Computational Intelligence Magazine}, vol. 4, no. 3, pp.
  62--76, 2009.

\bibitem{LSGO_Expensive_02}
Dominique Douguet,
\newblock ``{e-LEA3D}: a computational-aided drug design web server,''
\newblock {\em Nucleic acids research}, vol. 38, no. suppl\_2, pp. 615--621,
  2010.

\bibitem{LSGO_Expensive_03}
Soroush Rashidi and Prakash Ranjitkar,
\newblock ``Bus dwell time modeling using gene expression programming,''
\newblock {\em Computer-Aided Civil and Infrastructure Engineering}, vol. 30,
  no. 6, pp. 478--489, 2015.

\bibitem{Coordinate_Strategy}
Hans-Paul~Paul Schwefel,
\newblock {\em Evolution and optimum seeking: the sixth generation},
\newblock John Wiley \& Sons, Inc., 1993.

\bibitem{Adaptive_coordinate}
Ilya Loshchilov, Marc Schoenauer, and Mich{\`e}le Sebag,
\newblock ``Adaptive coordinate descent,''
\newblock in {\em Proceedings of the 13th annual conference on Genetic and
  evolutionary computation}, 2011, pp. 885--892.

\bibitem{Coordinate_CC}
Yi~Mei, Mohammad~Nabi Omidvar, Xiaodong Li, and Xin Yao,
\newblock ``A competitive divide-and-conquer algorithm for unconstrained
  large-scale black-box optimization,''
\newblock {\em ACM Transactions on Mathematical Software (TOMS)}, vol. 42, no.
  2, pp. 1--24, 2016.

\bibitem{center_sampling_01}
Shahryar Rahnamayan and G~Gary Wang,
\newblock ``Center-based sampling for population-based algorithms,''
\newblock in {\em IEEE Congress on Evolutionary Computation}. IEEE, 2009, pp.
  933--938.

\bibitem{DE_Deceptive_ICCSE2019}
Seyed~Jalaleddin Mousavirad, Azam Asilian~Bidgoli, and Shahryar Rahnamayan,
\newblock ``Tackling deceptive optimization problems using opposition-based
  {DE} with center-based {L}atin hypercube initialization,''
\newblock in {\em 14th International Conference on Computer Science and
  Education}, 2019.

\bibitem{center_sampling_03}
Hanan Hiba, Mohammed El-Abd, and Shahryar Rahnamayan,
\newblock ``Improving {SHADE} with center-based mutation for large-scale
  optimization,''
\newblock in {\em IEEE Congress on Evolutionary Computation}. IEEE, 2019, pp.
  1533--1540.

\bibitem{center_sampling_CEC2020}
Seyed~Jalaleddin Mousavirad and Shahryar Rahnamayan,
\newblock ``A novel center-based differential evolution algorithm,''
\newblock in {\em Congress on Evolutionary Computation}, 2020.

\bibitem{center_sampling_PSO_SMC2020}
Seyed~Jalaleddin Mousavirad and Shahryar Rahnamayan,
\newblock ``Cenpso: A novel center-based particle swarm optimization algorithm
  for large-scale optimization,''
\newblock in {\em International Conference on Systems, Man, and Cybernetics
  (IEEE SMC 2020)}, 2020.

\bibitem{CC_delta}
Mohammad~Nabi Omidvar, Xiaodong Li, and Xin Yao,
\newblock ``Cooperative co-evolution with delta grouping for large scale
  non-separable function optimization,''
\newblock in {\em IEEE Congress on Evolutionary Computation}. IEEE, 2010, pp.
  1--8.

\bibitem{CEC2010}
Tang Ki, Li~Xiaodong, PN~Suganthan, Zhenyu Yang, and Thomas Weise,
\newblock ``Benchmark functions for the {CEC2010} special session and
  competition on large-scale global optimization,''
\newblock {\em University of Science and Technology of China, Hefei, Anhui,
  China, and RMIT University, Australia, and Nanyang Technological University,
  Singapore Technical Report}, 2010.

\bibitem{CEC2017}
Guohua Wu, R~Mallipeddi, and PN~Suganthan,
\newblock ``Problem definitions and evaluation criteria for the {CEC 2017}
  competition on constrained real-parameter optimization,''
\newblock Tech. {R}ep., Nanyang Technological University, Singapore, 2016.

\end{thebibliography}

\end{document}